\documentclass[11pt]{article}

\usepackage[final]{acl}

\usepackage{times}
\usepackage{latexsym}
\usepackage{amsmath}
\usepackage{hyperref}
\usepackage{url}
\usepackage{algorithm}
\usepackage{algpseudocode}
\usepackage{booktabs}
\usepackage{multirow}
\usepackage{multicol}
\usepackage{footnote}
\makesavenoteenv{figure}

\usepackage[most]{tcolorbox}
\tcbuselibrary{listings} 
\definecolor{deepblue}{RGB}{0,51,102} 

\newtcolorbox{promptbox}[1][]{
  colback=gray!5,             
  colframe=black!40,          
  colbacktitle=deepblue,      
  coltitle=white,             
  fonttitle=\bfseries,        
  title=#1,                   
  breakable,                  
  enhanced,                   
  sharp corners,
  boxrule=0.5pt
}
\usepackage[T1]{fontenc}

\usepackage[utf8]{inputenc}

\usepackage{microtype}

\usepackage{inconsolata}

\usepackage{graphicx}

\title{On Improving Faithfulness of Podcasts from Documents}

\author{
 \textbf{Soumya Dutta\textsuperscript{1,2}},
 \textbf{Tejas Indulal Dhamecha\textsuperscript{2}},
 \textbf{Pannaga Shivaswamy\textsuperscript{2}}
\\
 \textsuperscript{1}Indian Institute of Science,
 \textsuperscript{2}Adobe Research
\\
 \small{
 \href{mailto:soumyad@adobe.com}{soumyad@adobe.com}
 }
}

\begin{document}
\maketitle
\begin{abstract}
Large language models (LLMs) are increasingly used to generate long-form conversational content such as podcasts from textual sources. While these systems produce fluent and engaging narratives, they often introduce ungrounded information. In this work, we present the first systematic study of faithfulness in document-grounded podcast generation, where grounding must be maintained across conversational turns in long-form, multi-speaker transcripts. We construct a dataset of over $1{,}500$ documents spanning five domains and generate podcast transcripts using multiple LLMs. We introduce a turn-level LLM-as-a-judge framework for evaluating whether conversational turns are supported by the source document, and validate its reliability through human studies. Our analysis shows that even state-of-the-art models, including \texttt{GPT-4o}, frequently generate ungrounded content. To mitigate this issue, we propose \textit{catch-n-repair}, a model-agnostic framework that detects and rewrites unfaithful conversational turns while preserving conversational flow. Experiments demonstrate consistent improvements in faithfulness across both in-domain and out-of-domain settings.
\end{abstract}
\section{Introduction}\label{sec:intro}

Podcasts have emerged as one of the most popular means of sharing information, spanning domains such as news reporting, education, entertainment, and lifestyle. A recent Pew Research Center report highlights this growth: while only $12\%$ of Americans aged $12$ and older listened to a podcast in a typical month in $2013$, the proportion rose to $42\%$ by $2023$~\citep{pew2023audio}. This surge in adoption reflects not only the familiarity of audiences with the medium but also the increasing trust placed in podcasts as a source of information. Until recently, however, podcasting largely remained a medium of consumption, where listeners tuned in to content created and curated by human hosts. The paradigm is now shifting with the advent of generative AI systems, such as Google’s NotebookLM\footnote{\url{https://notebooklm.google.com/}}, which can produce podcast-style narratives directly from source materials. This shift opens up the possibility of personalized podcast creation at scale, where individuals and organizations can generate podcasts tailored to their own needs and interests.
\begin{figure}[t]
    \centering
    \includegraphics[trim=5cm 11cm 2cm 3cm, clip, width=0.45\textwidth]{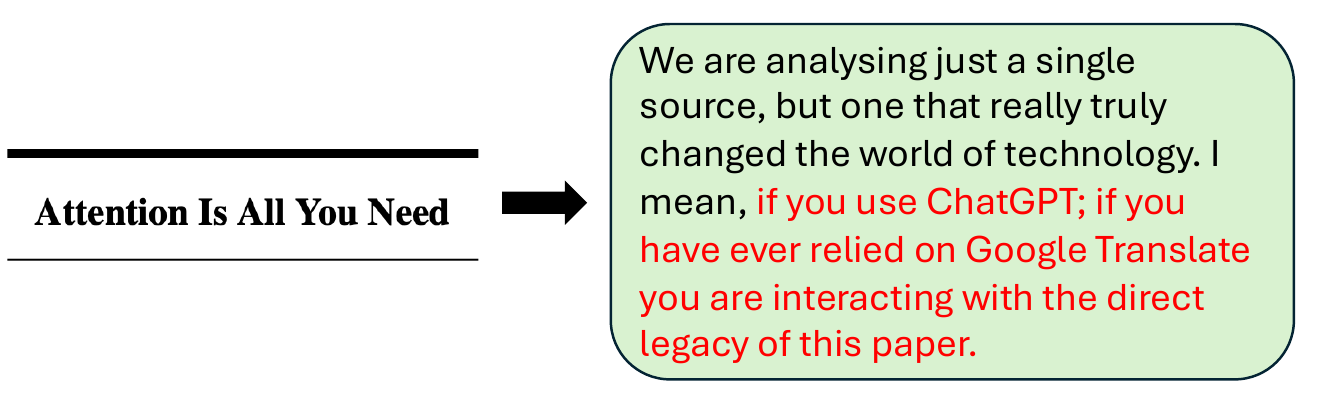}
    \caption{Example of unfaithfulness in generated podcast transcripts. The transcript generated from ‘Attention Is All You Need’ paper mentions ChatGPT which did not exist in 2017.}
    \label{fig:teaser}
\end{figure}

This development builds on a broader line of work in generative AI applications such as summarization~\citep{zhang2024benchmarking,kim-kim-2025-nexussum,liu2025sara} and question answering (QA)\citep{zaib2022conversational,hu-etal-2025-beyond,zhang-etal-2025-belle}, where models transform source material into more accessible forms. A central challenge in these tasks is \emph{faithfulness}—ensuring that generated outputs remain grounded in the source without introducing distortions or hallucinations~\citep{chiesurin2023dangers,jia2023zero,ming2024faitheval}.

While AI-generated podcasts inherit these concerns, they introduce a distinct set of challenges. Unlike summaries or answers, podcasts are typically long-form, narrative-driven, and multi-speaker, requiring models to maintain coherence and factual accuracy across extended spans~\citep{xiao-etal-2025-podagent}. Moreover, faithfulness in this setting must be preserved at the level of individual turns, where hallucinations can be localized yet propagate through the narrative. Despite recent efforts in podcast generation~\citep{ju2025mooncast,xiao-etal-2025-podagent,pengvibevoice}, prior work has primarily focused on fluency, structure, and audio delivery, leaving the question of grounding to the source document largely unexplored.

To illustrate this challenge, consider the example in Figure~\ref{fig:teaser}. The source is the Attention Is All You Need paper~\footnote{\url{https://arxiv.org/pdf/1706.03762}}
. \texttt{NotebookLM} generated a transcript mentioning ChatGPT as part of the paper’s legacy; while true in hindsight, ChatGPT did not exist in 2017 and there is no mention of ChatGPT in the source paper making this an unfaithful statement. This example highlights a subtle but critical distinction between factual correctness and faithfulness to the provided context. 

Motivated by this observation, we undertake the first systematic study of faithfulness in \emph{document-grounded podcast generation}. In contrast to prior work, we focus on long-form, multi-speaker transcripts and analyze faithfulness at the level of individual conversational turns. To enable this study, we collect a dataset of $1520$ documents spanning five domains and generate corresponding podcast transcripts using multiple large language models (LLMs). We introduce a \textbf{turn-level LLM-as-a-judge evaluation protocol} that assesses the degree to which each conversational turn is supported by the source document. This approach captures nuanced cases such as partial grounding and subtle hallucinations, which are common in long-form generation. We further validate the reliability of this evaluation framework through human studies.
  Building on this evaluation framework, we propose \textit{catch-n-repair}, a simple yet effective, model-agnostic mitigation strategy. Our approach operates at the level of individual conversational turns: it detects unfaithful generations using a lightweight classifier and rewrites them to ensure grounding in the source document while preserving conversational flow. Importantly, \textit{catch-n-repair} can be applied to both open-source and closed-source LLMs, making it broadly applicable in practical settings.
Our contributions are summarized as follows:
\begin{itemize}
\item \textbf{Evaluation protocol:} We introduce a turn-level LLM-as-a-judge framework for measuring faithfulness in long-form, multi-speaker podcast transcripts. The protocol captures fine-grained grounding at the level of individual conversational turns and is validated against human annotations.
\item \textbf{Benchmarking and analysis:} We conduct a systematic empirical study across multiple LLMs and domains, providing the first analysis of faithfulness in document-grounded podcast generation.
\item \textbf{Mitigation method:} We propose \textit{catch-n-repair}, a model-agnostic strategy that detects unfaithful conversational turns and rewrites them to ensure grounding in the source document, improving faithfulness while preserving conversational flow.

\end{itemize}

\section{Related Work}\label{sec:related}
\begin{figure*}[!ht]
    \centering
    \includegraphics[trim=1cm 2cm 1cm 9cm, clip, width=\textwidth]{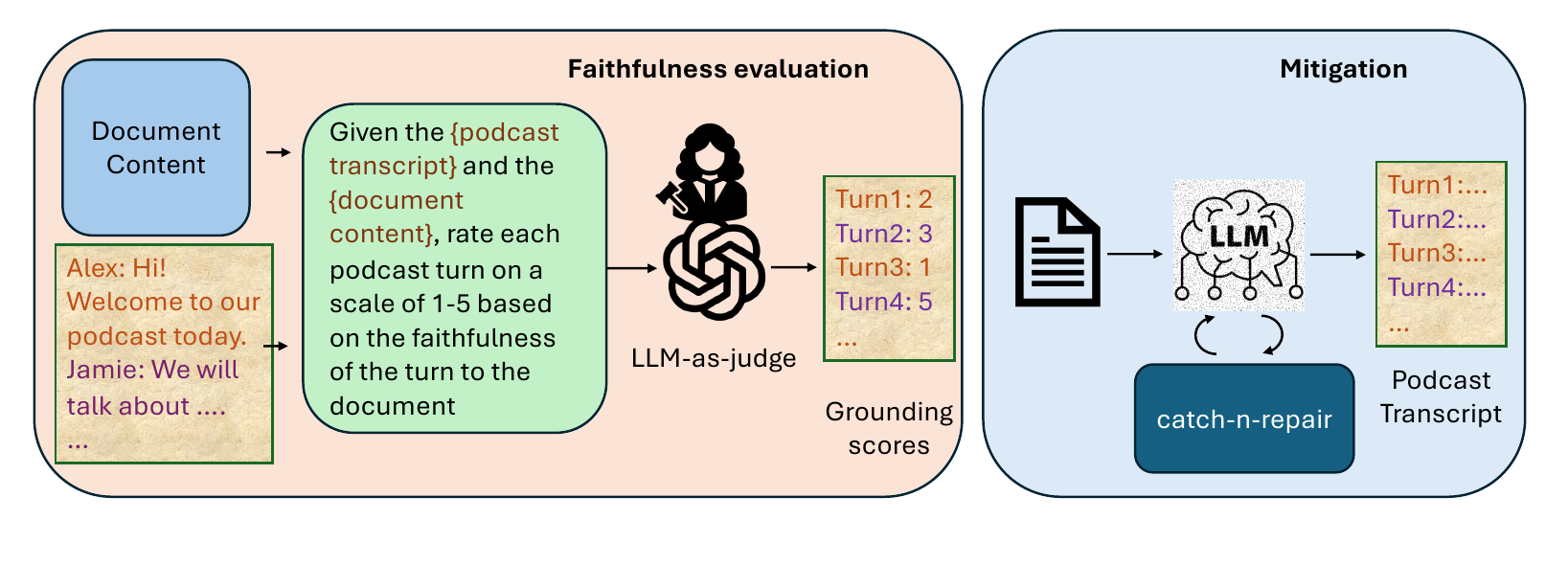}
    \caption{\textbf{Faithfulness evaluation}: Once these transcripts are generated, LLM-as-a-judge (\texttt{GPT4o}) evaluation is used to classify each turn of the podcast on a scale of $1$ to $5$ based on its faithfulness to the source document. \textbf{Mitigation}: The proposed mitigation strategy - \textit{catch-n-repair} - is used in conjunction with the LLM to generate transcripts that are faithful to the source document.}
    \label{fig:overall}
\end{figure*}

\textbf{Generation of Podcasts from Documents}: The task of generating podcasts directly from source documents has received relatively little attention in the research community. \citet{ju2025mooncast} proposed Mooncast, which leverages existing podcast scripts to produce long-form, engaging audio content. \citet{xiao-etal-2025-podagent} proposed PodAgent, an LLM agent that generates podcast scripts based on a user-specified topic, without relying on a source document. In contrast, our work focuses on \emph{document-grounded podcast generation} and systematically studies the faithfulness of generated transcripts to the underlying source material. \\
\textbf{Faithfulness in LLM Generation}: Ensuring that model outputs remain faithful to the provided context has been widely studied in tasks such as summarization and question answering~\citep{chiesurin2023dangers,jia2023zero,ming2024faitheval}.
Existing metrics, including token-overlap measures such as Knowledge-F1~\citep{shuster2021retrieval}, and learned evaluators such as FaithCritic~\citep{dziri2022faithdial}, have been proposed to detect hallucinations. More recently, LLM-as-a-judge frameworks~\citep{zheng2023judging,adlakha2024evaluating} have demonstrated strong performance in evaluating correctness and faithfulness. However, prior work has largely focused on short-form generation, where outputs are typically single-turn and concise. In contrast, podcast transcripts are long-form, multi-speaker, and narrative-driven, requiring faithfulness to be maintained at the level of individual conversational turns. Our work extends the study of faithfulness to this more complex setting.\\
\textbf{Evaluation of Podcast Transcripts}: Recent work has begun to explore the evaluation of podcast-style outputs generated by LLMs, often using LLM-as-a-judge frameworks to assess properties such as coherence, fluency, and engagement~\cite{xiao-etal-2025-podagent, xu2026podbench}. However, these approaches primarily operate at the level of the entire transcript and do not explicitly evaluate grounding with respect to a source document. As a result, they are not designed to detect unfaithful content that may arise during generation. In contrast, our work focuses on \emph{faithfulness to the source document} as a primary evaluation objective and introduces a turn-level evaluation protocol that enables fine-grained assessment of grounding within long-form, multi-speaker transcripts.\\
\textbf{Mitigating Ungroundedness in LLMs}: A wide range of approaches have been proposed to mitigate hallucinations in LLM outputs, where models generate content that is either factually incorrect or not grounded in the provided context. In this work, we focus on the latter—\emph{ungroundedness} with respect to a source document. Notably, ungrounded statements may be factually correct in the real world but remain unsupported by the input document, making them particularly challenging to detect in long-form generation.\\
Existing approaches mitigate hallucinations from open-weight LLMs through several directions. Decoding-based approaches adjust token probabilities to reduce ungrounded generations~\citep{shi-etal-2024-trusting,huangyw-etal-2025-dynamic}. Other methods perform post-hoc correction using auxiliary models or knowledge graphs~\citep{dziri2021neural,ji2023rho}. However, many of these approaches require access to model internals (e.g., logits), limiting their applicability in black-box settings. For such settings, prior work has largely relied on external knowledge or self-consistency signals. For instance, Ji et al.~\cite{ji2023towards} formulate validation as a question-answering task over domain-specific knowledge bases, while Manakul et al.~\cite{manakul2023selfcheckgpt} detect hallucinations through inconsistencies across multiple sampled generations. Gao et al.~\cite{gao2023rarr} further propose correcting outputs by retrieving supporting evidence from the web.\\
In contrast, our proposed \textit{catch-n-repair} framework is model-agnostic and operates at the level of individual conversational turns, enabling faithful generation without requiring access to model internals or external retrieval systems.

\section{Problem Statement}\label{sec:method}

\subsection{Problem Setup}\label{sec:prob_desc}

Let $\mathcal{D} = {d_1, d_2, \dots, d_n}$ denote a collection of source documents. For each document $d_i \in \mathcal{D}$, a large language model (LLM), denoted by $\mathcal{L}_{\text{gen}}$, generates a podcast-style conversational transcript:
$[
\mathcal{T}_i = {t_{i1}, t_{i2}, \dots, t_{im_i}}
]$
where each $t_{ij}$ represents the $j^\text{th}$ conversational turn in the transcript.

Our objective is to evaluate and improve the faithfulness of generated podcast transcripts with respect to their source documents. To this end, we define a faithfulness evaluation function $\mathcal{F}_{\text{eval}}$ that assigns a faithfulness score to each conversational turn $t_{ij}$ conditioned on the source document $d_i$.
The overall faithfulness score $f_i$ for transcript $\mathcal{T}_i$ is obtained by averaging across its $m_i$ turns:  
\begin{equation}\label{eq:faith_score}
    f_i = \frac{1}{m_i} \sum_{j=1}^{m_i} \mathcal{F}_{\text{eval}}(t_{ij}, d_i).
\end{equation}
Unlike prior settings that evaluate grounding at the response or document level, we formulate faithfulness at the level of individual conversational turns. This granularity is particularly important in long-form podcast generation, where ungrounded content is often localized to specific turns while the remainder of the transcript remains faithful to the source. Moreover, turn-level assessment enables targeted mitigation without requiring regeneration of the entire transcript.

Our goal is to design a mitigation method, \textit{catch-n-repair}, which can be applied to the generating model $\mathcal{L}_{\text{gen}}$ in order to increase the faithfulness score; i.e. to obtain improved transcript $\mathcal{T}^\prime_i$ such that the corresponding faithfulness score improves $f^\prime_i>f
_i$. An overview of the proposed framework is shown in Figure~\ref{fig:overall}.


\subsection{Creation of Podcast Transcripts\label{sec:create}}
We consider a podcast generation setup in which transcripts are generated sequentially in a turn-by-turn manner. Specifically, we focus on a two-speaker conversational structure commonly observed in real-world podcasts, where a host guides the discussion and an expert provides detailed, document-grounded responses.
The host turns are expected to contain natural and engaging questions that facilitate conversational flow, while the expert turns are expected to remain faithful to the source document. The transcript generation process is driven by an LLM $\mathcal{L}_{\text{gen}}$, whose sequential nature enables turn-level monitoring and intervention during generation.
\subsection{Evaluating Faithfulness of Generated Podcast Transcripts} \label{sec:faithfeval}
Designing the faithfulness scoring function $\mathcal{F}_{\text{eval}}$ is non-trivial due to two key challenges. First, conversational turns often express information through paraphrasing, making lexical overlap insufficient for evaluating grounding. Pure token-overlap metrics may therefore incorrectly penalize semantically faithful generations.  Second, the LLM $\mathcal{L}_{\text{gen}}$, may introduce content drawn from 
its parametric memory rather than from $d_i$ itself. Such hallucinated statements can be factually correct in the real world yet remain ungrounded with respect to the provided document.

To address these challenges, we adopt an \textit{LLM-as-a-judge} framework for computing $\mathcal{F}_{\text{eval}}$. Given a source document $d_i$ and a transcript turn $t_{ij}$,
an LLM evaluator model $\mathcal{L}_{\text{eval}}$ is prompted to assess the degree to which the factual content of $t_{ij}$ is supported by $d_i$. Rather than issuing a binary verdict, $\mathcal{L}_{\text{eval}}$ assigns a \emph{Likert-style} score $r_{ij}\in \{1,2,3,4,5\}$, where $1$ indicates that the turn is entirely unfaithful and $5$ indicates a faithful generation. This score directly defines $\mathcal{F}_{\text{eval}}(t_{ij}, d_i)=r_{ij}$, and the document-level faithfulness $f_i$ is computed as the mean of the turn-level scores (see Eq.~\ref{eq:faith_score}).
Using a graded scale allows the evaluator to capture nuanced cases, such as partially faithful turns or minor omissions, while remaining sensitive to hallucinated content. The evaluation prompt instructs the model to assess whether the response is explicitly grounded in the provided context, penalizing plausible but unsupported claims.\\
\textbf{Coverage Score}: While faithfulness captures the reliability of generated transcripts, podcast usability also depends on how comprehensively the transcript reflects the key content of the source document. We therefore additionally evaluate \emph{coverage}, which measures the extent to which important information from $d_i$ is represented in $\mathcal{T}_i$.
Unlike faithfulness, which is evaluated at the level of individual turns, coverage is assessed at the transcript level. The evaluator model $\mathcal{L}_{\text{eval}}$ is prompted to assign a score indicating how comprehensively the generated transcript covers the major points contained in the source document. Although we report coverage scores for completeness, improving coverage is not the primary focus of this work and is left as an important direction for future research.
\begin{figure*}[t]
    \centering
    \includegraphics[trim=1cm 4.5cm 1cm 2cm, clip, width=\textwidth]{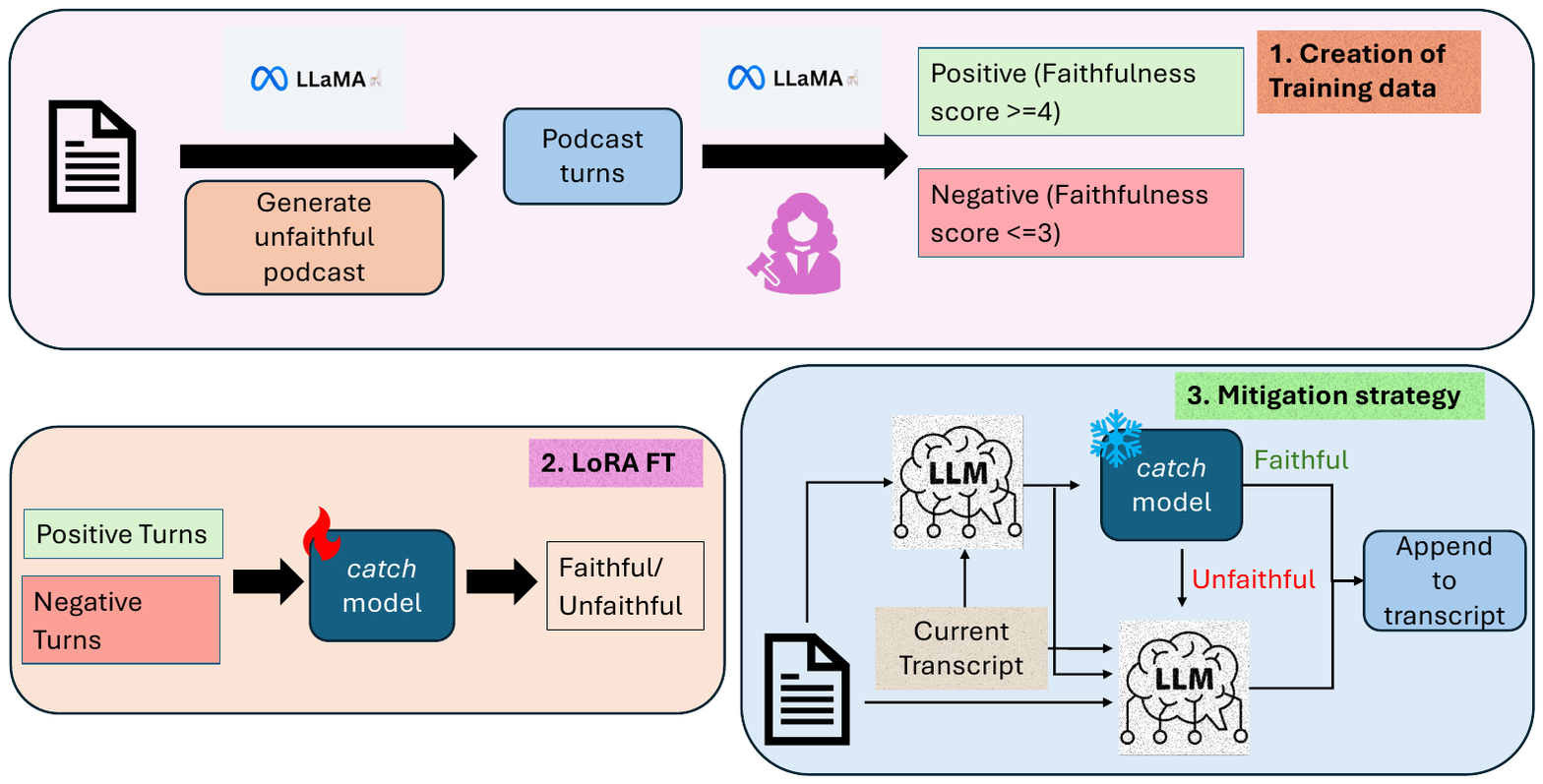}
    \caption{\textbf{Step 1}: Using an LLM to create training data for the \textit{catch} model. \textbf{Step 2}: Training another LLM for predicting faithful/unfaithful given any podcast turn. \textbf{Step 3}: The \textit{catch-n-repair} mechanism for mitigating the unfaithful generations. The \textit{catch} model flags each generation as faithful or unfaithful. The unfaithful generations are replaced by the same LLM with the old unfaithful generation as a negative example.}
    \label{fig:cnr}
\end{figure*}
\section{Catch-n-Repair} \label{sec:mitigate}
We now present \textit{catch-n-repair}, our proposed method for mitigating unfaithful generations in podcast transcripts. The design of this strategy is guided by two key goals: (\textit{i}) applicability to both open-source and closed-source models, and
(\textit{ii}) improvement of the faithfulness score $f_i$ with minimal impact on the coverage score, as evaluated by $\mathcal{L}_{\text{eval}}$ (Sec.~\ref{sec:faithfeval}).
To achieve these goals, we train an auxiliary model that approximates the judgment of $\mathcal{L}_{\text{eval}}$ and deploy it during transcript generation to identify unfaithful conversational turns.
\subsection{Training Data} \label{sec:data}

To train the detection component of \textit{catch-n-repair}, we construct a synthetic dataset of grounded and ungrounded conversational turns. For this stage, we employ the \texttt{LLaMA-3.1-70B} model as both the generation and evaluation model. Using an open-weight model ensures reproducibility and avoids practical and legal restrictions associated with creating supervision data using proprietary APIs.
Importantly, the resulting mitigation framework is evaluated using a different evaluator model.\\
To generate ungrounded examples, we modify the generation instructions to explicitly encourage the model to introduce information beyond the source document. Specifically, the model is encouraged to draw from general world knowledge, make tangential associations, and introduce plausible but unsupported details.\\
The resulting transcript $\mathcal{T}_i^{\text{hallu}}$ is then evaluated using the Likert-scale protocol described in Sec.~\ref{sec:faithfeval}. Turns receiving a faithfulness score $\geq 4$ are labeled \emph{faithful} (grounded), while all others are labeled \emph{unfaithful} (ungrounded), yielding a binary classification dataset for training the detection component of \textit{catch-n-repair}. This results in $17,561$ \textit{faithful} turns and $9,436$ \textit{unfaithful} turns.

\subsection{Training the \textit{catch} model} \label{sec:train}
With the synthetic dataset constructed in Sec.~\ref{sec:data},
we now describe how to train the model that detects unfaithful generations from $\mathcal{L}_{\text{gen}}$. Each training instance is a tuple $(t^{\text{hallu}}_{ij}, d_i, l_{ij})$, where $t^{\text{hallu}}_{ij}$ is a hallucinated turn,
$d_i$ is the source document, and $l_{ij} \in \{0,1\}$ denotes whether the turn is faithful.

Because this is a binary classification task, we initially considered encoder-only architectures.
However, modern encoders such as ModernBERT~\cite{warner-etal-2025-smarter}
still have a maximum context length of only $8192$ tokens,
while many source documents $d_i$ in our dataset are substantially longer.
To accommodate these long contexts,
we instead perform instruction fine-tuning of a large language model,
specifically \texttt{LLaMA-3.1-8B},
using LoRA~\cite{hu2022lora} to predict either ``faithful'' or ``unfaithful.''
The model is trained with an instruction format that presents the context document and the conversational response, then asks whether the response is faithful to the provided context.
For efficiency, the context document $d_i$ is truncated so that the complete instruction
does not exceed $32{,}768$ tokens. We further balance the training dataset between the two classes while training the \textit{catch} model. Once trained, this model can be seamlessly integrated into the generation pipeline to flag unfaithful turns.

\subsection{Repair Strategy}\label{sec:repair}
We now describe the repair component of \textit{catch-n-repair}. Our goal is to correct unfaithful generations while maintaining applicability to both open-source and black-box LLMs. This requirement rules out approaches that depend on access to model internals, such as activation steering~\cite{turner2023steering,rimsky2024steering} or confidence-based decoding strategies~\cite{bi2025parameters,liu2025selfelicit}.

While the \textit{catch} model is trained to predict whether each generated turn is faithful, our repair mechanism follows a simple but effective strategy. Whenever a conversational turn $t_{ij}$ is flagged as unfaithful, $\mathcal{L}_{\text{gen}}$ is re-prompted to revise $t_{ij}$ to produce a corrected turn that is semantically faithful to the provided context while preserving conversational flow.
The regeneration prompt instructs the model to rewrite the turn $t_{ij}$ using only facts explicitly stated in the source document, while maintaining natural conversational flow. Importantly, the repair process is integrated into the sequential generation pipeline, allowing unfaithful turns to be corrected before subsequent turns are generated. This reduces propagation of ungrounded content across the conversation. The repaired turn is then appended directly to the transcript. The final transcript produced by $\mathcal{L}_{\text{gen}}$ under the \textit{catch-n-repair} mitigation strategy is denoted by $\mathcal{T}_{i}^{\text{faith}}$. The workflow is shown in Fig.~\ref{fig:cnr}.
\section{Experiments}\label{sec:experiments}

\subsection{Doc-to-Podcast}\label{sec:dataset}
We collected a total of $1500$ documents to create our \textit{Doc-to-Podcast} dataset. The corpus spans five domains—\textit{academic}, \textit{legal}, \textit{policy}, \textit{financial}, and \textit{medical}—with $300$ documents sampled from each domain.
Further details are in Appendix~\ref{sec:dataset_Details}.

To capture a broad range of input complexities, we deliberately varied the length of the collected documents. The distribution of document sizes across domains is shown in Appendix Fig.~\ref{fig:dist}; all documents contain between $2$ and $50$ pages. The dataset serves two purposes: (i) it is used to evaluate the faithfulness of podcasts generated by multiple LLMs ($\mathcal{L}_{\text{gen}}$), and (ii) it provides training and validation material for our proposed \textit{catch} model.

For testing the \textit{catch-n-repair} mitigation strategy, we use $40$ held-out documents. Of these, $20$ are from the \textit{Doc-to-Podcast} dataset ($4$ per domain), forming the \emph{in-domain} test set.  
The remaining $20$ documents originate from unrelated areas such as \textit{exam evaluations} and \textit{business practices}, and constitute the \emph{out-of-domain} test set.

\begin{figure}[t]
    \centering
    \includegraphics[trim=3cm 8cm 5cm 4cm, clip, width=0.5\textwidth]{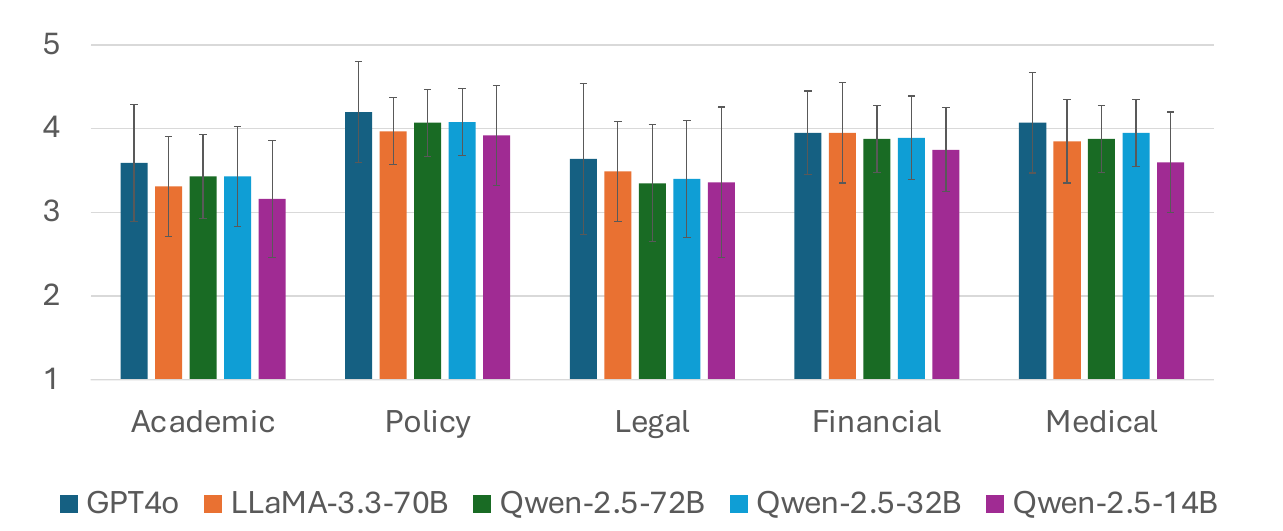}
    \caption{The average faithfulness score as evaluated with $\mathcal{L}_{\text{eval}}$ set as \texttt{GPT-4o}. The scores for different $\mathcal{L}_{\text{gen}}$ are shown for the different domains.}
    \label{fig:faitheval}
\end{figure}

\subsection{Faithfulness across LLMs}\label{sec:faitheval}
\begin{figure}[t]
    \centering
    \includegraphics[trim=4cm 5cm 4cm 4cm, clip, width=0.48\textwidth]{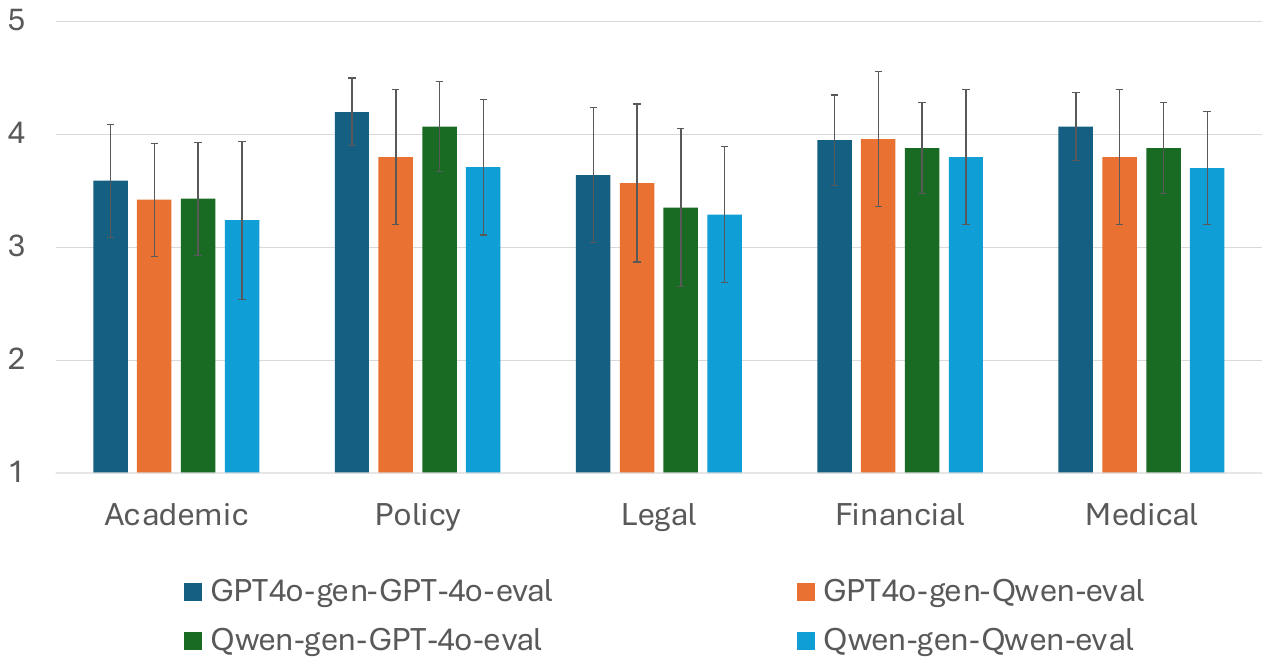}
    \caption{The average faithfulness score as evaluated with the Qwen model when the generating LLM is either Qwen or GPT-4o. The evaluations with GPT-4o model are also shown for reference.}
    \label{fig:qwen}
\end{figure}

Figure~\ref{fig:faitheval} shows the average faithfulness scores across all documents in each domain for different generating LLMs. All podcast turns are evaluated by \texttt{GPT-4o} acting as the judge LLM. For transcript generation, we use three variants from the \texttt{Qwen2.5} family ($14$B, $32$B, and $72$B), the $70$B model from the \texttt{LLaMA3.3} family, and \texttt{GPT-4o} itself. The \texttt{LLaMA-3.1-70B} model is not used in this case in order to avoid overlap with the training paradigm of the \textit{catch} model (Sec.~\ref{sec:data}).
\begin{itemize}
    \item \textbf{GPT-4o achieves the highest faithfulness}: Among all models, transcripts generated by \texttt{GPT-4o} receive the highest average faithfulness score of $3.89$ on the Likert scale, followed by \texttt{Qwen2.5-32B} with a score of $3.75$. Notably, even the strongest model fails to consistently achieve fully grounded generation, indicating that maintaining faithfulness remains challenging in long-form conversational settings despite substantial advances in LLM capabilities.
    \item \textbf{Bigger open models aren’t automatically more faithful}: While increasing size generally helps, our results show diminishing returns beyond the mid-scale range. For instance, \texttt{LLaMA3.3-70B} performs on par with \texttt{Qwen2.5-72B} and \texttt{Qwen2.5-32B}, while the smaller \texttt{Qwen2.5-14B} lags behind at $3.55$.
    \item \textbf{Faithfulness varies by domain}: Scores are not uniform across domains. The academic domain records the lowest average faithfulness ($3.38$ across all models), while the policy domain achieves the highest ($4.05$). This suggests that the intrinsic knowledge distribution of LLMs influences their ability to remain faithful to the document.
    \item \textbf{Cross-LLM agreement alleviates bias concerns}: To further assess evaluator reliability, we additionally replace \texttt{GPT-4o} with \texttt{Qwen2.5-72B} as the evaluator model. The results mirrored those of \texttt{GPT-4o}: \texttt{Qwen} also rated \texttt{GPT-4o}’s generations as more faithful than its own (Fig.~\ref{fig:qwen}). This agreement across independent evaluator models reduces concerns regarding model-specific judge bias.
\end{itemize}
We present the coverage scores and observations from them in Appendix~\ref{sec:coverage}.

\subsection{Reliability of the Evaluator}\label{ref:llmeval}
In order to assess the reliability of our evaluator, we conduct a human annotation study using $30$ randomly selected documents ($6$ from each domain) from the \textit{Doc-to-Podcast} dataset and annotate a total of $328$ conversational turns. The annotations were collected from $10$ annotators (Sec.~\ref{sec:annot_desc}). We measure inter-annotator agreement using Krippendorff’s $\alpha$, obtaining a value of $0.69$, indicating substantial agreement. We compare the scores assigned by our evaluator model (\texttt{GPT-4o}) against human ratings, and also benchmark the evaluator against existing automatic faithfulness metrics. K-Prec., K-Rec., and K-F1 compute lexical overlap between the podcast turn and the source document~\cite{shuster2021retrieval}, while K-BERTs-F1 and K-BERTs-Prec. extend this approach using BERTScore-based semantic similarity~\cite{adlakha2024evaluating,zhang2019bertscore}.\\
The Pearson correlations for each domain are shown in Table~\ref{tab:faith_eval_compare}. \texttt{GPT-4o} achieves the highest correlation with human judgments across all domains, with an average correlation of $0.63$. In contrast, existing automatic faithfulness metrics exhibit substantially weaker alignment. These results suggest that LLM-based evaluation provides a more reliable proxy for human assessment in long-form conversational generation. Interestingly, model-free lexical grounding metrics such as $K$-Precision consistently outperform learned evaluators such as \textit{FaithCritic}, with the exception of \texttt{GPT-4o}. This observation is consistent with prior findings reported by Adlakha et al.~\citep{adlakha2024evaluating}. 
Notably, these correlations are obtained in a more challenging setting than conventional short-form evaluation.
\begin{table}[t]
\centering
\setlength{\tabcolsep}{3.5pt}
\renewcommand{\arraystretch}{1.1}
\resizebox{\columnwidth}{!}{%
\begin{tabular}{l|ccccc|c}
\toprule
\textbf{Metric} & \textbf{Acad.} & \textbf{Legal} & \textbf{Fin.} & \textbf{Policy} & \textbf{Med.} & \textbf{Avg.} \\
\midrule
K-Prec. & $\underline{0.44}$ & $\underline{0.48}$ & $0.24$ & $\underline{0.54}$ & $0.47$ & $\underline{0.43}$ \\
K-Rec.  & $0.11$ & $0.18$ & $0.13$ & $0.10$ & $0.19$ & $0.14$ \\
K-F1    & $0.11$ & $0.18$ & $0.13$ & $0.10$ & $0.19$ & $0.14$ \\
\midrule
K-BERTs-F1 & $0.02$ & $0.19$ & $\underline{0.26}$ & $0.42$ & $\underline{0.53}$ & $0.28$ \\
K-BERTs-Prec. & $-0.13$ & $0.02$ & $0.20$ & $0.19$ & $0.51$ & $0.16$ \\
FaithCritic & $0.10$ & $0.19$ & $0.16$ & $0.02$ & $0.29$ & $0.15$ \\
\midrule
GPT-4o & $\mathbf{0.65}$ & $\mathbf{0.61}$ & $\mathbf{0.59}$ & $\mathbf{0.68}$ & $\mathbf{0.64}$ & $\mathbf{0.63}$ \\
\bottomrule
\end{tabular}}

\caption{Pearson correlation with human faithfulness scores across domains. GPT-4o shows consistently higher alignment than existing metrics.}
\label{tab:faith_eval_compare}
\end{table}
\begin{table*}[t]
\centering
\small
\setlength{\tabcolsep}{4pt}
\renewcommand{\arraystretch}{}
\begin{tabular}{p{0.40\linewidth}|p{0.25\linewidth}|p{0.25\linewidth}}
\toprule
\textbf{Source Document Summary} & \textbf{Vanilla Generation} & \textbf{After \textit{catch-n-repair}} \\
\midrule

The source document is an environmental science coursework handout focused on topics such as BPA exposure, recycling, environmental health, and sustainability. It contains educational explanations, reflective discussion prompts, and assessment-style questions intended to encourage students to think critically about environmental risks and personal choices related to health and pollution.

&
Environmental science education seems like such a critical topic these days. Can you explain \textbf{why understanding human impacts on the environment is so important?}

&
I've been reading about environmental science education focusing on human impacts. It's fascinating how it integrates assessments and activities related to pollutants and toxic chemicals. Could you share more about the types of assessments or activities mentioned in the document?

\\
\bottomrule
\end{tabular}
\caption{Example of correcting an ungrounded conversational turn using \textit{catch-n-repair}.}
\label{tab:qualitative}
\end{table*}
\subsection{Training and Evaluation of the \textit{catch} Model}\label{sec:catch}

\textbf{Implementation Details}: We fine-tune a \texttt{LLaMA-3.1-8B} model to classify each transcript turn as \textit{faithful} or \textit{unfaithful} given the source document as context. To respect computational limits, each document is truncated so that the combined input remains below $32,768$ tokens. Fine-tuning is performed with a batch size of $8$ and gradient accumulation of $8$ using LoRA~\cite{hu2022lora} (rank = $8$, $\alpha = 16$, dropout = $0.05$). Training uses $90\%$ of the documents from the synthetic dataset (Sec.~\ref{sec:data}), while turns from the remaining $10\%$ form the validation set.\\
\textbf{Performance}: The validation split contains $3{,}039$ turns, including $1{,}973$ faithful and $1{,}066$ unfaithful examples. The trained \textit{catch} model achieves F1-scores of $79.9\%$ and $67.3\%$ on the faithful and unfaithful classes, respectively. The lower performance on unfaithful examples reflects the difficulty of detecting subtle yet plausible ungrounded content. Despite being substantially smaller than the original evaluator (\texttt{LLaMA-3.1-70B}), the resulting $8$B model remains lightweight and suitable for runtime deployment.

\subsection{Effectiveness of \textit{catch-n-repair}}\label{sec:results}
Table \ref{tab:faith_results} presents the faithfulness scores for both in-domain (ID) and out-of-domain (OOD) settings, comparing vanilla generation with our \textit{catch-n-repair} (CnR) strategy. 
\begin{itemize}
    \item \textbf{\textit{catch-n-repair} consistently boosts faithfulness}: Across all models and both ID and OOD settings, CnR improves faithfulness scores. 
    The largest relative gain for the ID test set is seen for the \texttt{Qwen2.5-32B} and \texttt{Qwen2.5-72B} models ($\sim7.5\%$). For the OOD test set, the largest gain of $12.8\%$ is seen for the \texttt{LLaMA-3.3-70B} model.
    \item \textbf{Robust to domain shift}: Faithfulness gains hold for the OOD test set. On average, across all models, the relative increase in faithfulness scores are more for the OOD set ($\sim10\%$) versus the ID set ($\sim5\%$).
    \item \textbf{\texttt{GPT-4o} remains the overall leader}: \texttt{GPT-4o} achieves the highest faithfulness in both ID and OOD settings, with an average relative increase of $7.4\%$ after the mitigation strategy. The consistent gains obtained from \textit{catch-n-repair} indicate that even state-of-the-art proprietary models continue to benefit from explicit turn-level grounding interventions.
\end{itemize}

\begin{table}[t]

\setlength{\tabcolsep}{3.5pt}
\renewcommand{\arraystretch}{1.1}
\resizebox{\columnwidth}{!}{%
\begin{tabular}{l|ccc|ccc|c}
\toprule
\multirow{2}{*}{$\mathcal{L}_{\text{gen}}$} &
  \multicolumn{3}{c|}{ID} &
  \multicolumn{3}{c|}{OOD} &
  \multirow{2}{*}{Avg. $\Delta$} \\
\cmidrule(lr){2-4}\cmidrule(lr){5-7}
  & Van. & CnR & $\Delta$ & Van. & CnR & $\Delta$ &  \\
\midrule
\texttt{Qwen2.5-14B}  & $4.1^{\pm0.5}$ & $4.2^{\pm0.6}$ & $+0.1$ & $3.7^{\pm0.7}$ & $4.1^{\pm0.5}$ & $+0.4$ & $+0.25$ \\
\texttt{Qwen2.5-32B}  & $4.1^{\pm0.5}$ & $\mathbf{4.4}^{\pm0.5}$ & $+0.3$ & $\mathbf{4.0}^{\pm0.6}$ & $4.2^{\pm0.5}$ & $+0.2$ & $+0.25$ \\
\texttt{Qwen2.5-72B}  & $4.0^{\pm0.4}$ & $4.3^{\pm0.4}$ & $+0.3$ & $3.9^{\pm0.3}$ & $4.3^{\pm0.4}$ & $+0.4$ & $\mathbf{+0.35}$ \\
\texttt{LLaMA3.3-70B} & $\mathbf{4.2}^{\pm0.5}$ & $4.3^{\pm0.5}$ & $+0.1$ & $3.9^{\pm0.5}$ & $\mathbf{4.4}^{\pm0.4}$ & $+0.5$ & $+0.30$ \\
\texttt{GPT-4o}       & $\mathbf{4.2}^{\pm0.3}$ & $\mathbf{4.4}^{\pm0.3}$ & $+0.2$ & $\mathbf{4.0}^{\pm0.4}$ & $\mathbf{4.4}^{\pm0.3}$ & $+0.4$ & $+0.30$ \\
\bottomrule
\end{tabular}}
\caption{Faithfulness scores for in-domain (ID) and out-of-domain (OOD) test sets. 
Van. denotes vanilla prompting (Sec.~\ref{sec:create}), CnR indicates the \textit{catch-n-repair} strategy, 
$\Delta$ represents the absolute improvement, and Avg. $\Delta$ is the mean improvement across ID and OOD.}
\label{tab:faith_results}
\end{table}

\paragraph{Qualitative Analysis:} Table~\ref{tab:qualitative} presents an example of conversational turns before and after applying \textit{catch-n-repair}. The source document primarily consists of environmental science coursework and assessment instructions related to BPA exposure, recycling, and environmental health. In the vanilla generation setting, the conversation deviates from discussing the specific content in the document to broader discussions about the importance of assessing the impact that humans have on the environment. While these additions are plausible, they are not explicitly grounded in the source document. After applying \textit{catch-n-repair}, the generated response adheres more closely to the scope and content of the original document.

The performance of CnR with \texttt{Qwen-2.5-72B} as judge is shown in Appendex~\ref{sec:cnrqwen}. The impact of CnR on coverage is shown in Appendix~\ref{sec:cnrcoverage}. The performance of the \textit{catch} model with a text classifier instead of a LLM is shown in Sec.~\ref{sec:llmcatch}.


\section{Conclusion}
We present the first study of faithfulness in document-grounded podcast generation, which involves long-form, multi-speaker conversations. Our analysis across multiple LLMs and domains shows that even state-of-the-art models frequently generate ungrounded content, highlighting the difficulty of maintaining faithfulness in long-form conversational generation.
To address this, we introduce a turn-level LLM-as-a-judge evaluation framework and validate its reliability against human judgments. We further propose \textit{catch-n-repair}, a model-agnostic mitigation strategy that detects and corrects unfaithful conversational turns. Experiments demonstrate consistent improvements in faithfulness across both in-domain and out-of-domain settings.
Overall, our work highlights faithfulness as an important challenge in AI-generated podcasts and demonstrates the effectiveness of turn-level grounding mechanisms for trustworthy conversation generation systems.
\section{Limitations}
While our work provides a first benchmark for faithfulness in document-to-podcast generation, several limitations remain. First, our evaluation relies primarily on English-language documents; extending the dataset to multilingual and culturally diverse sources is an important future step. Second, although our \textit{catch-n-repair} method improves faithfulness, it depends on the accuracy of the \textit{catch} classifier, which may occasionally mislabel subtle paraphrases as unfaithful. Third, our faithfulness assessment is based on prompts that were designed for this task. Different prompts might result in different results. Another limitation of our work is the potential misuse, even after our attempts some ungrounded generations may still occur.
Finally, the podcasts in our dataset are text-based transcripts rather than full audio renderings — future work could explore faithfulness in end-to-end audio generation pipelines, where prosody and delivery interact with factual grounding.

\bibliography{custom}

\appendix
\newpage
\section{Appendix}

\subsection{Dataset details}\label{sec:dataset_Details}
\begin{figure}[ht]
    \centering
    \includegraphics[trim=3cm 7.5cm 7cm 4cm, clip, width=0.5\textwidth]{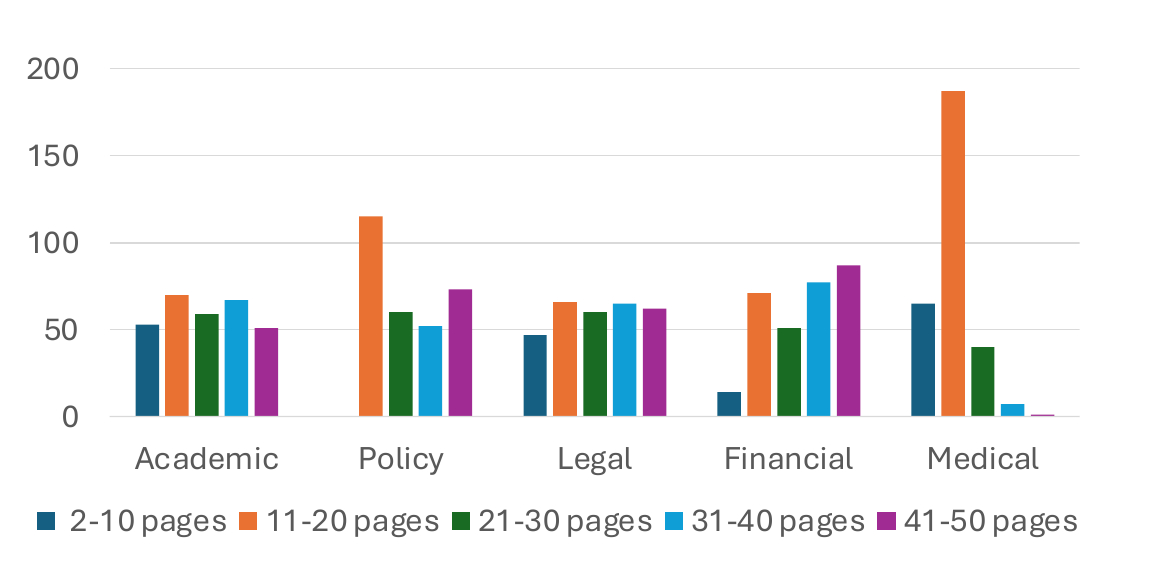}
    \caption{The number of documents in each page range for the different domains.}
    \label{fig:dist}
\end{figure}
We provide the links from which the documents are collected for each of the domains below:
\begin{itemize}
    \item Academic: \url{https://arxiv.org/}
    \item Policy: \url{https://gov-report-data.github.io/}
    \item Legal: \url{https://indiankanoon.org/}
    \item Financial: \url{https://www.annualreports.com/Companies?ind=i5}
    \item Medical: \url{https://pmc.ncbi.nlm.nih.gov/}
\end{itemize}

\subsection{Evaluator Robustness Across Judge Models}\label{sec:cnrqwen}

One potential concern in LLM-as-a-judge evaluation is whether the observed improvements are specific to a particular evaluator model. To assess the robustness of our findings, we repeat the evaluation of \textit{catch-n-repair} using \texttt{Qwen2.5-72B} as the evaluator in place of \texttt{GPT-4o}.\\
The resulting faithfulness scores are shown in Table~\ref{tab:cnr_qwen}. Across all generator models and both in-domain (ID) and out-of-domain (OOD) settings, \textit{catch-n-repair} continues to improve faithfulness scores relative to vanilla generation. Moreover, the overall ranking of generator models remains broadly consistent across evaluators.\\
While the absolute scores differ slightly between evaluators, the qualitative trends remain stable. In particular, both evaluators consistently identify improvements in faithfulness after applying \textit{catch-n-repair}, suggesting that the observed gains are not artifacts of a specific judge model. These results reduce concerns regarding evaluator-specific bias and support the robustness of our conclusions across independent LLM evaluators.

\begin{table}[t]

\setlength{\tabcolsep}{3.5pt}
\renewcommand{\arraystretch}{1.1}
\resizebox{\columnwidth}{!}{%
\begin{tabular}{l|ccc|ccc|c}
\toprule
\multirow{2}{*}{$\mathcal{L}_{\text{gen}}$} &
  \multicolumn{3}{c|}{ID} &
  \multicolumn{3}{c|}{OOD} &
  \multirow{2}{*}{Avg. $\Delta$} \\
\cmidrule(lr){2-4}\cmidrule(lr){5-7}
  & Van. & CnR & $\Delta$ & Van. & CnR & $\Delta$ &  \\
\midrule
\texttt{Qwen2.5-14B}  & $3.9^{\pm0.3}$ & $4.2^{\pm0.3}$ & $+0.3$ & $3.5^{\pm0.5}$ & $4.0^{\pm0.4}$ & $+0.4$ & $+0.35$ \\
\texttt{Qwen2.5-32B}  & $4.0^{\pm0.2}$ & $4.3^{\pm0.4}$ & $+0.3$ & $3.8^{\pm0.4}$ & $4.1^{\pm0.4}$ & $+0.2$ & $+0.25$ \\
\texttt{Qwen2.5-72B}  & $3.8^{\pm0.5}$ & $4.3^{\pm0.3}$ & $+0.5$ & $\mathbf{4.0}^{\pm0.4}$ & $4.2^{\pm0.3}$ & $+0.4$ & $+0.45$ \\
\texttt{LLaMA3.3-70B} & $3.9^{\pm0.3}$ & $4.2^{\pm0.2}$ & $+0.3$ & $3.7^{\pm0.4}$ & $4.3^{\pm0.3}$ & $+0.6$ & $+0.45$ \\
\texttt{GPT-4o}       & $\mathbf{4.1}^{\pm0.2}$ & $\mathbf{4.5}^{\pm0.4}$ & $+0.4$ & $3.9^{\pm0.3}$ & $\mathbf{4.4}^{\pm0.3}$ & $+0.5$ & $+0.45$ \\
\bottomrule
\end{tabular}}
\caption{Faithfulness scores for in-domain (ID) and out-of-domain (OOD) test sets. 
Van. denotes vanilla prompting (Sec.~\ref{sec:create}), CnR indicates the \textit{catch-n-repair} strategy, 
$\Delta$ represents the absolute improvement, and Avg. $\Delta$ is the mean improvement across ID and OOD. The judge is \texttt{Qwen-2.5-72B}.}
\label{tab:cnr_qwen}
\end{table}
\begin{table*}[t]
\centering
\setlength{\tabcolsep}{4pt}
\begin{tabular}{l|cc|cc|cc|cc}
\toprule
\multirow{3}{*}{$\mathcal{L}_{\text{gen}}$} &
  \multicolumn{4}{c|}{In-Domain (ID)} &
  \multicolumn{4}{c}{Out-of-Domain (OOD)} \\
\cmidrule(lr){2-5}\cmidrule(lr){6-9}
  & \multicolumn{2}{c|}{Faithfulness} & \multicolumn{2}{c|}{Coverage}
  & \multicolumn{2}{c|}{Faithfulness} & \multicolumn{2}{c}{Coverage} \\
\cmidrule(lr){2-3}\cmidrule(lr){4-5}\cmidrule(lr){6-7}\cmidrule(lr){8-9}
  & Van. & CnR & Van. & CnR & Van. & CnR & Van. & CnR \\
\midrule
\texttt{Qwen2.5-14B}  & $4.1^{\pm0.5}$ & $4.2^{\pm0.6}$ & $3.8^{\pm0.4}$ & $3.5^{\pm0.7}$ & $3.7^{\pm0.7}$ & $4.1^{\pm0.5}$ & $3.4^{\pm0.5}$ & $3.3^{\pm0.5}$ \\
\texttt{Qwen2.5-32B}  & $4.1^{\pm0.5}$  & $\mathbf{4.4}^{\pm0.5}$  & $4.0^{\pm0.6}$  & $4.0^{\pm0.5}$  & $\mathbf{4.0}^{\pm0.6}$  & $4.2^{\pm0.5}$  & $4.0^{\pm0.4}$  & $4.0^{\pm0.5}$  \\
\texttt{Qwen2.5-72B}  & $4.0^{\pm0.4}$  & $4.3^{\pm0.4}$  & $4.3^{\pm0.6}$  & $4.1^{\pm0.7}$  & $3.9^{\pm0.3}$  & $4.3^{\pm0.4}$  & $4.3^{\pm0.4}$  & $4.1^{\pm0.7}$  \\
\texttt{LLaMA3.3-70B} & $\mathbf{4.2}^{\pm0.5}$  & $4.3^{\pm0.5}$  & $2.8^{\pm0.6}$  & $3.1^{\pm0.4}$  & $3.9^{\pm0.5}$  & $\mathbf{4.4}^{\pm0.4}$  & $2.5^{\pm0.5}$  & $2.8^{\pm0.6}$  \\
\texttt{GPT-4o}       & $\mathbf{4.2}^{\pm0.3}$ & $\mathbf{4.4}^{\pm0.3}$  & $\mathbf{4.9}^{\pm0.4}$  & $\mathbf{4.8}^{\pm0.4}$  & $\mathbf{4.0}^{\pm0.4}$  & $\mathbf{4.4}^{\pm0.3}$  & $\mathbf{4.7}^{\pm0.5}$  & $\mathbf{4.6}^{\pm0.5}$  \\
\bottomrule
\end{tabular}
\caption{Faithfulness and coverage for in-domain (ID) and out-of-domain (OOD) test sets.
Van. stands when the LLMs are prompted as described in Sec.~\ref{sec:create}, while CnR represents the scores when unfaithfulness is mitigated using the \textit{catch-n-repair} strategy.}
\label{tab:faith_cov_results}
\end{table*}
\subsection{Evaluating Coverage in \textit{Doc-to-Podcast}}\label{sec:coverage}
\begin{figure}[ht]
    \centering
    \includegraphics[trim=2cm 8cm 4cm 4cm, clip, width=0.5\textwidth]{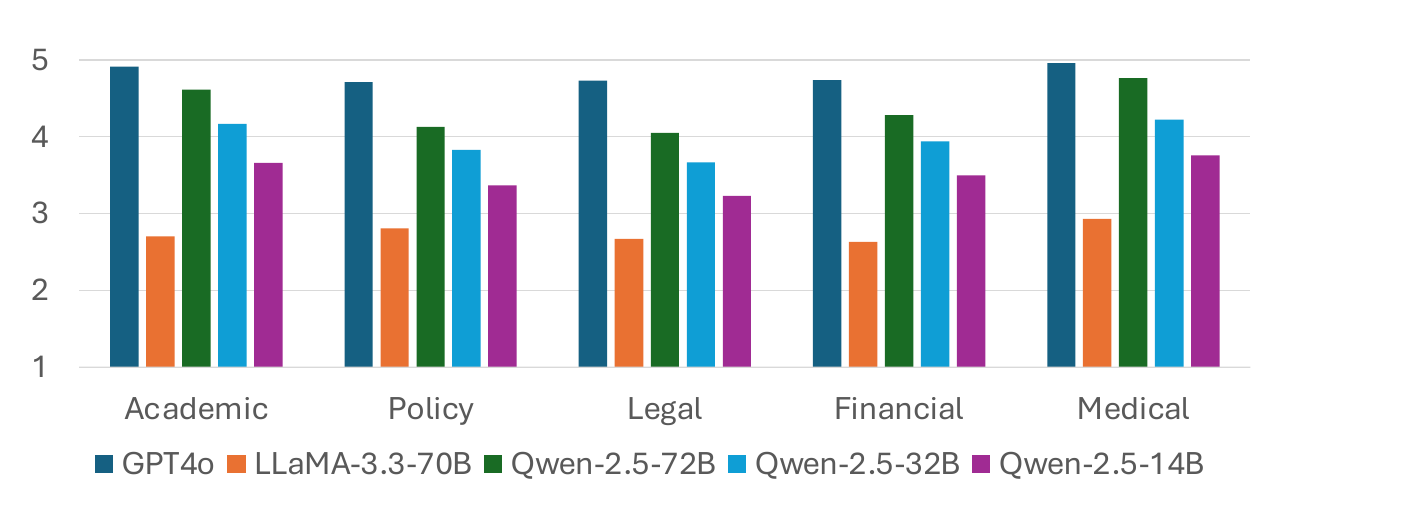}
    \caption{The average coverage score as evaluated with $\mathcal{L}_{\text{eval}}$ set as \texttt{GPT-4o}. The scores for different $\mathcal{L}_{\text{gen}}$ are shown for the different domains.}
    \label{fig:coverage}
\end{figure}

The coverage scores for the different domains and the different generating LLMs are shown in Fig.~\ref{fig:coverage}. 
\begin{itemize}
    \item \textbf{GPT-4o achieves the highest coverage}: Among all models, transcripts generated by \texttt{GPT-4o} receive the highest average coverage score of $4.81$ on the Likert scale, followed by \texttt{Qwen2.5-72B} with a score of $4.37$.
    \item \textbf{Coverage scores of LLaMA are strikingly low}: The coverage scores assigned to \texttt{LLaMA} are strikingly low at $2.75$. The similar sized \texttt{Qwen2.5-72B} gets a score of $4.37$. This highlights a potential issue with using \texttt{LLaMA} as the generating LLM.
    \item \textbf{Bigger models are better in the Qwen family}: As the model size increases in the \texttt{Qwen2.5} family, the coverage score increases. While the $14$B model has an average coverage score of $3.5$, the $32$B model is much better at $3.97$. The $72$ B model is the best of the three, with an average coverage score of $4.37$.
    \item \textbf{Coverage varies by domain}: Like faithfulness, the coverage score also varies according to the domain. It is highest for the medical domain ($4.13$) and lowest for the legal domain ($3.67$)
\end{itemize}

\subsection{Impact of \textit{catch-n-repair} on Coverage}\label{sec:cnrcoverage}
The results for faithfulness and coverage are shown in Table~\ref{tab:faith_cov_results}. The following are the key takeaways.
\begin{itemize}
    \item \textbf{Minimal impact on coverage}: Coverage remains largely unchanged, with only small decreases ($\leq 0.2$ in most cases), indicating that the mitigation strategy does not significantly reduce transcript completeness. 
    \item \textbf{\texttt{GPT-4o is the best model for faithfulness and coverage}}: \texttt{GPT-4o} remains the best model in terms of balancing faithfulness and coverage in the generated podcasts.
\end{itemize}

\subsection{Are LLMs really required for the \textit{catch} model?}\label{sec:llmcatch}
The \textit{catch} component can be viewed as a text classification problem: given a podcast turn and its corresponding source document, the model predicts whether the turn is \textit{faithful} or \textit{unfaithful}. An attentive reader might question the need for using an LLM for this classification objective. To investigate this, we replace the \texttt{LLaMA-3.1-8B} backbone with the encoder-based \texttt{ModernBERT} model~\cite{warner-etal-2025-smarter}. We observe that the F1-score on faithful validation examples decreases from $79.9\%$ to $76.1\%$, while performance on unfaithful examples drops more substantially from $67.3\%$ to $60.9\%$. These results suggest that long-context LLMs provide a significant advantage for detecting subtle forms of ungroundedness in long-form conversational generation. In particular, identifying unsupported yet factually plausible statements often requires reasoning over large portions of the source document, which remains challenging for shorter-context encoder architectures. Since the effectiveness of the \textit{catch-n-repair} pipeline depends critically on accurately identifying unfaithful turns, stronger long-context reasoning substantially improves the reliability of downstream repair.

\subsection{License}

 \begin{itemize}
     \item \texttt{Qwen2.5 models} - Apache license (\url{https://huggingface.co/Qwen/Qwen2.5-14B-Instruct/blob/main/LICENSE})
     \item \texttt{LLaMA models} - LLaMA license (\url{https://www.llama.com/llama3/license/})
 \end{itemize}

 \subsection{Use of LLMs}

 LLMs were used for polishing the writing and also for help during coding.

\subsection{Human annotation setup}\label{sec:annot_desc}
The human evaluation study was conducted using $10$ student annotators recruited from the university community. Annotators were presented with a source document and a generated podcast turns from the system described in Sec.~\ref{sec:create}, and were asked to rate the faithfulness of the podcast turns with respect to the source document on a Likert scale from $1$ to $5$.

Annotators were specifically instructed to evaluate whether the information in the conversational turn was supported by the source document, rather than whether the content was factually plausible in general. In particular, they were asked to penalize unsupported extrapolations, policy-oriented additions, or conversational embellishments not grounded in the document. Each example was independently annotated by multiple annotators, and the final score was obtained by averaging the collected ratings. Annotators were financially compensated in accordance with local standards. The instruction for the human annotation is shown here.

\paragraph{Instruction to humans:}
Please go through the PDF. Grounding means explicit support from the context. Plausible or theoretically possible claims that are not mentioned in the context should be penalized. Please assign a score from 1--5 using the following guidelines: (1)
5(Completely Grounded): ALL claims/questions are explicitly supported by the context. No speculation or extrapolation beyond what is directly stated.
(2) 4(Mostly Grounded): Core claims are explicitly supported, with only very minor inferences that stay within the scope of the provided material.
(3) 3(Partially Grounded): Some connection to context but includes unsupported claims, applications, or examples not mentioned in the source material.
(4) 2(Mostly Ungrounded): Relates to the context but makes several significant unsupported claims or extrapolations beyond what is stated, even if plausible.
(5) 1(Completely Ungrounded): Contradicts the context directly OR is completely unrelated/irrelevant to the source material.



\end{document}